\documentclass[11pt,a4paper]{article}
\usepackage[hyperref]{naaclhlt2019}
\usepackage{times}
\usepackage{latexsym}

\usepackage{url}            %
\usepackage{booktabs}       %
\usepackage{amsfonts}       %
\usepackage{nicefrac}       %
\usepackage{microtype}      %
\usepackage{graphicx}       %
\usepackage{multirow}
\usepackage{tabularx} 
\usepackage{paralist}
\usepackage{dblfloatfix}

\aclfinalcopy %
\def\aclpaperid{16} %

\usepackage{color}

\newcommand{\E}{{\mathcal E}}

\newcommand{\R}{{\mathcal R}}

\newcommand{\tX}{\tilde{X}}
\newcommand{\tK}{\tilde{K}}

\usepackage{xspace}
\newcommand{\ttoQ}{{\sc Text2Quest}\xspace}
\newcommand{\tlabs}{{\sc TextLabs}\xspace}

\usepackage{color}
\newcommand{\dnote}[1]{\textcolor{blue}{$\ll$\textsf{#1 --Dafna}$\gg$}}

\newcommand{\remove}[1]{}

\title{Playing by the Book: An Interactive Game Approach for Action Graph Extraction from Text}

\author{Ronen Tamari\thanks{\ \ Work was begun while author was an intern at RIKEN and continued at the Hebrew University.} \\
  The Hebrew University of Jerusalem \\
  {\tt ronent@cs.huji.ac.il} \\\And
  Hiroyuki Shindo \\
  NAIST / RIKEN-AIP \\
  {\tt shindo@is.naist.jp}\\
  \AND
  Dafna Shahaf \\
  The Hebrew University of Jerusalem\\
  \tt{dshahaf@cs.huji.ac.il} \\
  \And
  Yuji Matsumoto \\
  NAIST / RIKEN-AIP \\
  \tt{matsu@is.naist.jp} \\
}
\date{}
\begin{document}
\maketitle

\begin{abstract}

  Understanding procedural text requires tracking entities, actions and effects as the narrative unfolds. We focus on the challenging real-world problem of action-graph extraction from {\it materials science} papers, where language is highly specialized and data annotation is expensive and scarce. We propose a novel approach, \ttoQ, where procedural text is interpreted as instructions for an \emph{interactive game}. A learning agent completes the game by executing the procedure correctly in a text-based simulated lab environment. The framework can complement existing approaches and enables richer forms of learning compared to static texts. We discuss potential limitations and advantages of the approach, and release a prototype proof-of-concept, hoping to encourage research in this direction. %
\end{abstract}

\section{Introduction}

\begin{figure*}[t!]
\centering
\includegraphics[width=.95\linewidth]{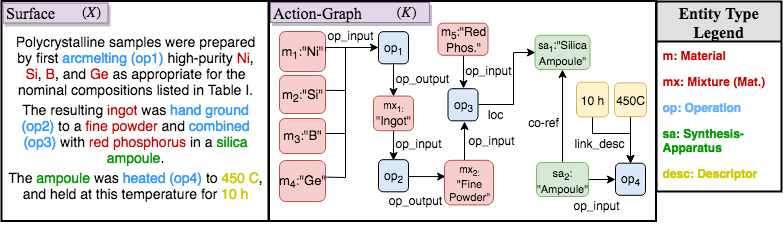}
\caption{\label{fig:example} Sample surface text (left) and possible corresponding action-graph (right) for typical partial material synthesis procedure. Operation numbers in parentheses are added for clarity. Nodes are entities, edges are relations linking them, equivalent to actions in the text-based game.}
\end{figure*}

Materials science literature includes a vast amount of synthesis procedures described in natural language. The ability to automatically parse these texts into a structured form could allow for data-driven synthesis planning, a key enabler in the design and discovery of novel materials \citep{Kim2018,Mysore2017}. A particularly useful  parsing is  \textbf{action graph extraction}, which maps a  passage describing a procedure to a symbolic action-graph representation of the core entities, operations and their accompanying arguments, as they unfold throughout the text  (Fig. \ref{fig:example}).

Procedural text understanding is a highly challenging task for today's learning algorithms \citep{Lucy2017,Levy2017}. Synthesis procedures are especially challenging, as they are written in difficult and highly technical language assuming prior knowledge. Some texts are long, many follow a non-linear narrative, or include logical quantifiers (``all synthesis steps were performed in an argon atmosphere...''). 
Furthermore, annotated data is scarce and expensive to obtain.

Two related research areas are \textbf{grounded semantic parsing} and \textbf{state-tracking reading-comprehension}. Grounded (or executable) semantic parsers map natural language to a symbolic representation which can also be thought of as a sequence of instructions in some pre-defined programming language. Such ``neural-programing'' architectures offer strong symbolic reasoning capabilities, compositionality modelling, and strong generalization \citep{Reed2015}, but are typically applied  to  simple texts due to prohibitive annotation costs \citep{Liang2016}. {\bf State-tracking} models  \citep{BosselutNPN2018, Das2018, Bansal2017} can model complex relations between entities as they unfold, with easier training but less symbolic reasoning abilities. Their applicability to longer texts is hindered as well by the lack of fine-grained annotated data.

\begin{figure}[!]
\centering
\includegraphics[width=.27\textwidth]{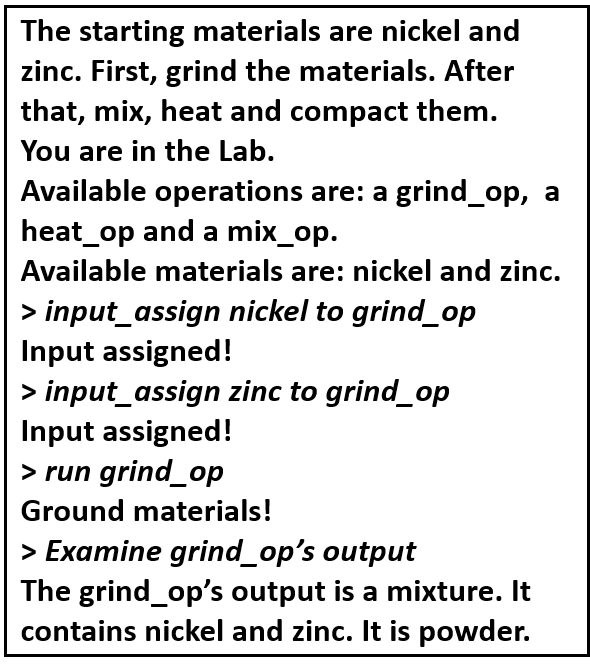}
\caption{\label{fig:tbg} Excerpt from an actual ``material synthesis quest'' generated by our system with example input/outputs.}
\end{figure}

In this work we describe an approach, \ttoQ, that attempts to combine the strengths of both methods. %
Instead of trying to learn from static text, we propose to treat procedural text as \textbf{instructions for an interactive game} (or ``quest''). The learning agent interacts with entities defined in the text by executing symbolic actions (Fig. \ref{fig:tbg}). A text-based symbolic interpreter handles execution and tracking of the agent's state and actions. The game is completed by ``simulating'' the instructions correctly; i.e., mapping instructions to a sequence of actions. Correct simulation thus directly yields the desired action graph.

While there is some engineering overhead required for the simulator, we demonstrate that it is relatively straightforward to convert an annotation schema to a text-based game. We believe that the benefits make it worth pursuing: the game format allows applying powerful neural programming methods, with a significantly richer training environment, including advances such as curriculum learning, common-sense and domain-specific constraints, and full state tracking.  Such ``friendly'' environments that assist the learning agent have been shown to be valuable \citep{Liang2016} and enable learning of patterns that are often hard to learn from surface annotations alone, such as implicit effects of operations (i.e., filtering a mixture splits it into two entities).
 
Interestingly, understanding by simulation aligns well with models of human cognition; mental simulation, the ability to construct and manipulate an internal world model, is a cornerstone of human intelligence involved in many unique behaviors, including language comprehension \citep{Marblestone2016TowardAI, Hamrick2019}.
In this work we take first steps towards this idea. Our contributions are: \remove{\dnote{Expand again}}
 
 \begin{compactitem}
 \item We propose a {\bf novel formulation} of the problem of procedural text understanding as a text-based game, enabling the use of neural programming and text-based reinforcement learning (RL) methods. 
 
 \item  We present and release \tlabs\footnote{Code and experiments available at \url{https://github.com/ronentk/TextLabs}}, an instance of \ttoQ designed for interaction with synthesis procedure texts. We focus on the material-science setting, but the approach is intended to be more generally applicable.
 
  \item We propose to address the problem of obtaining full-graph annotations at scale by coupling the simulator with \textbf{controllable natural language generation} (NLG) to generate synthetic data, also enabling curriculum learning. 

 \end{compactitem}

While this work is preliminary in nature, neural programming and text-based reinforcement learning approaches are attracting significant and growing interest, and we expect advances in these areas to directly benefit future versions of the system.

\section{Related Work}
\label{sec:related}

\remove{\dnote{this section is really missing some flow. Take the first line: you never say that the goal of process understanding is}}
\textbf{Procedure understanding}:  Many recent works have focused on tracking entities and relations in long texts, such as cooking recipes and scientific processes \citep{BosselutNPN2018,Das2018}. However, these methods do not directly extract a full action graph. For action graph extraction, earlier works use sequence tagging methods \citep{Mysore2017}. \citet{Feng2018} have applied deep-RL to the problem of extracting action sequences, but assume explicit procedural instruction texts. In \citet{Johnson2017LearningGS}, a graph is constructed from simple generated stories, using state tracking at each time step as supervision. 

\textbf{Semantic parsing \& Neural Programming}: Research to-date has focused mainly on shorter and simpler texts which may require complex symbolic reasoning, such as mapping natural language to queries over knowledge graphs \citep{Liang2016}. In the case of narrative parsing, the text itself may be complex while the programs are relatively simple (creating and linking between entities present in the text). Recent work \citep{Lu2018ObjectorientedNP} frames narrative understanding as neural-programming, the learner converts a document into a structured form, using a predefined set of data-structures. This approach is similar to ours, though with simpler texts and without a simulated environment. In our approach, the learning architecture is decoupled from the symbolic interpreter environment, enabling greater architectural flexibility.

\textbf{Text-RL}: Text-based games are used to study language grounding and understanding and RL for combinatorical action spaces \citep{Zahavy2018LearnWN,DBLP:phd/ndltd/Narasimhan17} but have not yet been applied to real world problems. TextWorld \citep{cote18textworld} is a recently released reinforcement learning sandbox environment for creation of custom text-based games, upon which we base \tlabs.

\section{Problem Formulation}
\textbf{Entities, Relations \& Rules ($\E, \R, \Lambda$)}: Assume two vocabularies defining types of \textit{entities} $\E=\left\{ e_{1},...,e_{N}\right\} $
and \textit{relations} $\R=\left\{ r_{1},...,r_{K}\right\}$. A \textit{fact}
$f$ is a grounded predicate of the form $f=r\left(h,t\right),\,h,t\in\E,\,r\in\R$ 
(single or double argument predicate relations are allowed). We define the set of valid world-states $S$, where a state $s \in S$ is a set of facts, and validity is decided by a world-model $\Lambda$ defined using linear logic. $\Lambda$ is comprised of production rules (or transition rules) over entities and relations governing which new facts can be produced from a given state. Following the schema used in the Synthesis Project\footnote{\url{https://www.synthesisproject.org/}} (see for example \citet{MSPT}), entity types include materials, operations, and relevant descriptors (like operation conditions, etc.). Relations link between entities (like \textit{input(material,operation)} or denote single predicate relations (entity properties such as \textit{solid(material)}). We currently use a simplified version of the schema to ease the learning problem. See appendix \ref{subsec:types} for a mapping of relations and entities. Production rules correspond to the actions available to the learner, in our domain these include for example \textit{link-descriptor(descriptor,entity)}, \textit{input-assign(material, operation)}. While not currently included, actions such as co-reference linking and generation of entities can also be incorporated.

\textbf{Action-Graph ($K$)}:
An action sequence is defined to be a sequence of valid actions (or production rules) rooted at some initial state $s_{0}$: $K=\left(s_{0},\lambda_{0},\lambda_{1},...,\lambda_{n}\right)$ (applying $\lambda_{i}$ to $s_{i}$ results in $s_{i+1}$, intermediate states are left out for brevity). Note that actions may apply to implicit entities not present in the surface text (for example, the result of an operation). Construction of an action graph corresponding to $K$ is straightforward (entities as nodes, actions connecting them as edges), and henceforth we use $K$ to denote either the sequence or the graph. Note that there can be multiple possible action sequences resulting in the same action graph, equivalent w.r.t the topological ordering of operations induced by their dependencies.

\textbf{Surface ($X$)}: A \textit{surface} is simply a text in natural language describing a process.

\textbf{Learning Task}:
Our objective is to learn a mapping $\Psi:X\to K$. As this mapping may be highly complex, we convert the problem to a structured prediction setting. As an intermediate step \textbf{we map an input $X$ to an enriched text-based-game $G$ representation} (details below), where the solution of $G$ is the required action graph $K$. The game is modelled as a partially observable Markov Decision Process (POMDP) $G=\left(S,A,T,\Omega,O,R,\gamma\right)$.

We refer the reader to \citet{cote18textworld} for a detailed exposition, and focus here on mapping the game-setting to our approach:
$S$ are states, $A$ are actions, $T$ are conditional state transition probabilities, where all are constant per domain and defined by $\E, \R, \Lambda$. $\Omega$ are observations, and $O$ are conditional observations probabilities. $R: S \times A \to \mathbb{R}$ is the reward function,  $\gamma \in \left[0,1\right]$ is the discount factor. 
As $\gamma, \Omega, O$  are also preset (with actual observations dependent on agent actions), mapping a surface $X$ to game $G$ boils down to providing a list of entities for initializing $s_0$. For training and evaluation, a reward function must also be provided (not necessary for applying a trained model on un-annotated text ``in the wild''). 

If a fully annotated action graph is available (whether synthetic or real), this mapping is simple: the initial game state $s_0$ is a room where the agent is placed alongside all entities. Each edge corresponds to an action in the game. Given an action sequence $K$, a reward function $R$ can be automatically computed, giving intermediate rewards and penalizing wrong actions. A quest in TextWorld can be defined via a final goal state, thus allowing multiple possible winning action sequences. See appendices \ref{subsec:quests-dets}, \ref{subsec:real-graphs} for examples\remove{from synthetic/real graphs}.

For data ``in the wild'', entities can be identified using named entity recognition (NER) as pre-processing. Future directions include end-to-end learning to reduce cascading initialization errors.\remove{Reward information is not available, though future research could explore intrinsic rewards  \citep{Singh2005}.}

By default, the TextWorld environment is partially observable. The agent observes the surface $X$ at time $t=0$ and other textual descriptions upon executing an ``examine'' action. Unlike classic text-based games where partial observability is part of the challenge, in our case we can adopt the ``friendly-environment'' perspective and assist the learner with information such as state-tracking or action pruning \remove{shown to be effective in}\citep{Liang2016, Johnson2017LearningGS}.

\begin{figure}[t!]
\centering
\includegraphics[width=.50\textwidth]{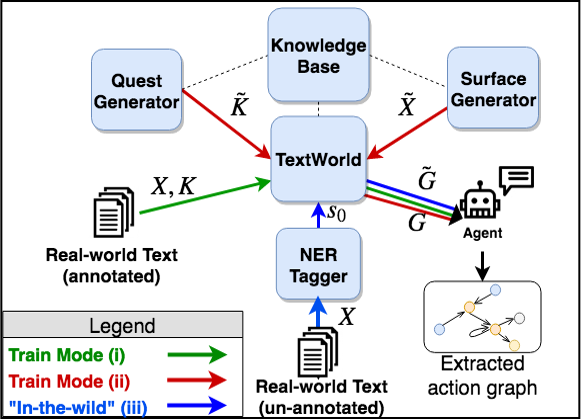}
\caption{\label{fig:arch} Proposed solution architecture of \ttoQ. (i) Flow for training agent on games from real annotated data. (ii) Flow for training agent on synthetic games. (iii) Extracting action graph from un-annotated real data.}
\end{figure}

\section{Proposed Solution Architecture}\label{sec:solution}
Our system consists of 6 core modules (Fig. \ref{fig:arch}): a Knowledge Base defines entity, relation and action vocabularies. This is used by the Surface Generator and Quest Generator modules to generate pairs ($\tX,\tK$) of synthetic surfaces and their corresponding action graphs for training. For un-annotated text, a pre-trained domain specific NER tagger\footnote{For the materials synthesis domain we use the tagger available at \url{https://github.com/olivettigroup/materials-synthesis-generative-models}} is used to extract an initial game state $s_{0}$ by identifying the mentioned entities. A learning agent extracts $K$ from a generated game.  

The \ttoQ architecture supports three central modes of operation: (i) Enrich existing real world annotated pairs ($X,K$) by converting them to game instances for training the game-solving agent. (ii)  Produce synthetic training pairs ($\tX,\tK$). (iii) Convert un-annotated texts to game instances for action graph extraction ``in the wild''.

The current version of \tlabs supports mode (ii). We implemented simple prototypes of the domain-specific Knowledge Base, plus Quest and Surface Generators. See Sec. \ref{subsec:types} for details about converting the entity and relation annotation schema into TextWorld. TextWorld is easily extensible and can support a variety of interaction semantics. Aside from adding a domain specific entity type-tree and actions, most of the underlying logic engine and interface is handled automatically\remove{\footnote{TextWorld is also still at preliminary stages, and some custom Inform7 code was required.}}. 
For the game environment, we use Inform7, a programming language and interpreter for text-based games.
\remove{Finally, it is intended to be a text-based games research platform, and not a framework for semantic parsers; purpose-built interpreters for this task could also reduce the environment engineering required.} For quest generation, we currently use simple forward chaining and heuristic search strategies to create plausible quests (for example, all start materials must be incorporated into the synthesis route). Combining these with a simple rule-based Surface Generator already allows for creating simple synthetic training game instances (Fig. \ref{fig:tbg}). 

\section{Preliminary Evaluation}
\label{sec:evaluation}
As a very preliminary sanity check for the \tlabs environment, we train a simple text-based RL agent on synthetic games in increasingly difficult environments. Difficulty is measured by maximum quest length, and the number of entities in the target action graph. See Sec. \ref{subsec:quests-dets} for representative examples. We use the basic LSTM-DQN agent of \citet{DBLP:phd/ndltd/Narasimhan17} adapted to the \tlabs setting. The action space is $A=\left\{ W_{v}\times W_{o_{1}}\times W_{o_{2}}\right\}$, where $W_{v}$ consists of 8 action-verbs corresponding to the entity relations tracked and additional native TextWorld actions like \textit{take} (see Sec. \ref{subsec:types} for details).  $W_{o_{1}},W_{o_{2}}$ are (identical) sets of potential arguments corresponding to the active entities which can be interacted with in the game (single and double argument actions allowed). As this basic agent is not conditioned on previous actions,  we further concatenate the last four commands taken to the current observation. For the same reason, we also append the full quest instructions at every timestep's observation. All illegal actions are pruned at each state to reduce search space size. 

We train the agent on 100 games per level and test on 10 games. Evaluation is measured by avg. normalized reward per game: $\frac{1}{\left|K\right|}\sum_{t=1}^{T}r_{i}$, where $K$ is the true action sequence, $T$ is the episode length (set to 50) and $r_i=1$ for each action in $K$ and $-1$ for otherwise (and 0 for neutral actions like \textit{examine}). A normalized score of 1 means the agent performed the required actions exactly.

\begin{figure}
  \includegraphics[width=1.05\columnwidth, trim={2cm 0 0 0},clip]{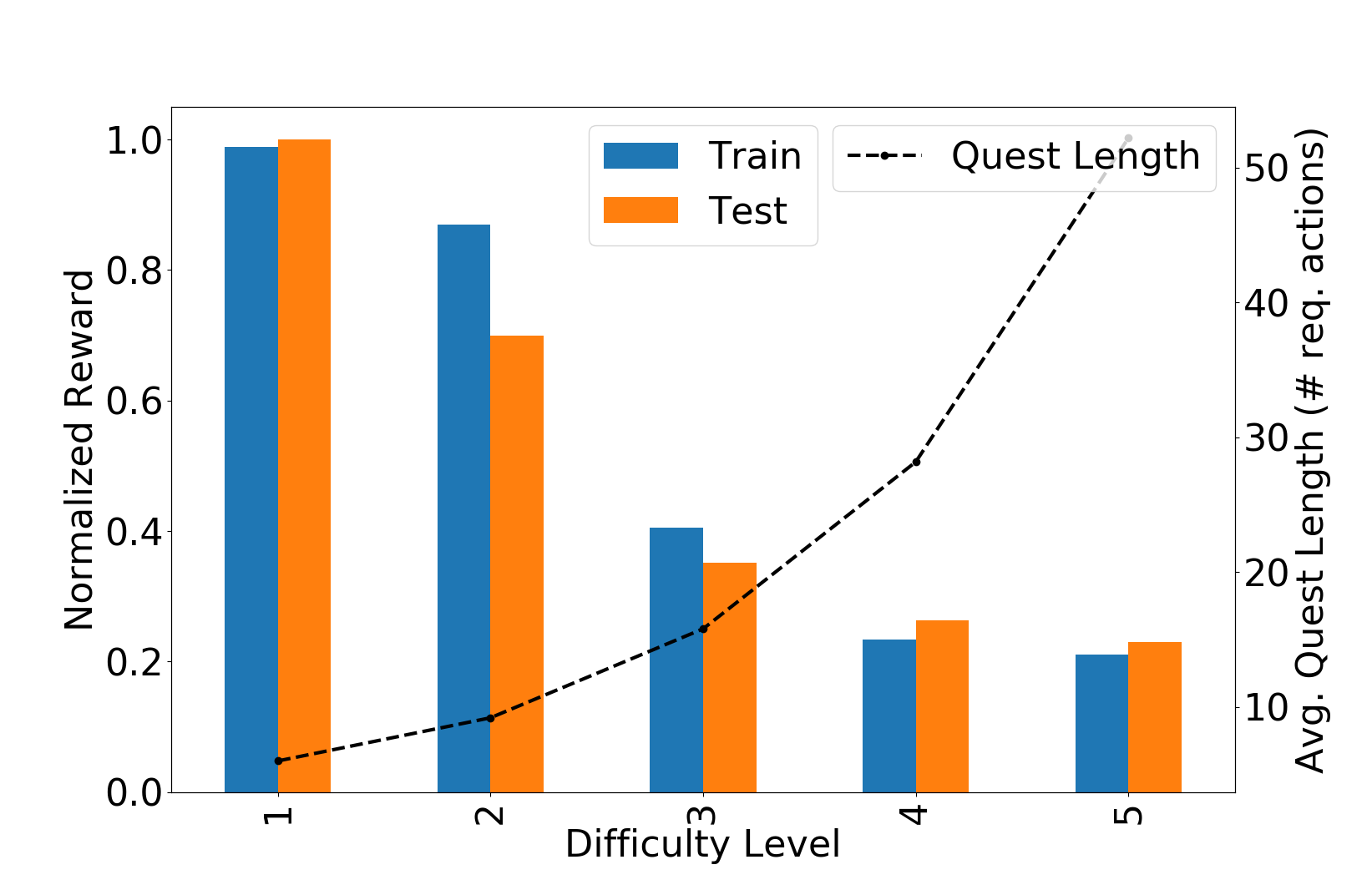}
  \caption{Preliminary evaluation results for a basic LSTM-DQN text-RL agent on synthetic quests. Dotted line shows average generated quest lengths. }
  \label{fig:eval}
\end{figure}

As can be seen in Fig. \ref{fig:eval}, the agent learns to successfully perform the required actions only for the easiest levels. Examining longer games the agent did not complete, we note that the lack of conditioning on previous states is a serious limitation. Equipping agents with better sequence encoding (e.g., attention), recurrent memory, and utilizing state information is expected to significantly improve performance. Furthermore, due to technical limitations of the current implementation, some actions cannot be reversed. This adds to the difficulty of the task, and will be addressed in future versions. Finally, learning good initial policies for semantic parsers is known to be a hard problem with RL alone, and related approaches commonly use hybrid RL/supervised training methods \citep{Liang2016,Jiang2012LearnedPF}.

\section{Discussion}\label{sec: discussion}
Our approach faces tough challenges. However, we are encouraged by the significant recent advances towards these challenges in related areas, and plan to leverage this progress for our framework.

Programming semantics and rewards for instruction-following agents is known to be notoriously difficult  \citep{WINOGRAD19721} as language and environments grow increasingly complex. Research on \textbf{learned instruction-conditional reward models} \citep{Bahdanau2018} is a promising approach towards reducing the amount of ``environment engineering'' required.
 
 Another critical open question in our framework is whether the surface generator will be able to generate surfaces representative enough to allow for generalization to real examples. Current NLG systems are increasingly capable of structured text generation \citep{Marcheggiani2018}, and though they produce relatively short surfaces, we believe that coupling them with the generated action graphs is a promising approach to scaling up to longer sequences while maintaining coherence. Such systems can use sentence-level semantic parses as training data, meaning they can leverage existing weakly-supervised shallow parsing techniques. Encouraging for our modelling paradigm, recent work \citep{Peng2018} extending the Dyna-Q (DQ) framework \citep{Sutton1990} demonstrates a real-world application of structured NLG with a simulated RL training environment.
 
 Given sufficient text generation capabilities, one may question the added utility of the game environment (as opposed to learning a direct mapping $X \to K$). Recent research suggests that for stronger generalization, data alone may not be enough, and symbolic reasoning capabilities are necessary \citep{Khashabi2018QuestionAA,Yi2018}. Given the compositional complexity and difficulty of the language involved, we believe they will prove necessary in our setting as well.

\section{Conclusions}
\label{sec:future}
There is a growing need for combining neuro-symbolic reasoning with advanced language representation methods\remove{stronger generalization and more complex reading comprehension capabilities}. In the case of procedural text understanding, key obstacles are suitable training environments, as well as the lack of fully annotated action graphs. Motivated by this, we proposed \ttoQ, an approach intended to enhance learning by turning raw text inputs into a structured \emph{text-based game} environment, as well as augmenting data with synthetic fully annotated action graphs. To encourage further research in this direction, we publicly release \tlabs, an instance of \ttoQ for the  materials synthesis task. We implemented prototype modules for basic game generation and solving. Future work will focus on designing learning agents to solve the games, as well as improving text generation capabilities.
We hope that the proposed approach will lead to developing useful systems for action graph extraction as well as other language understanding tasks.

\newpage 

\bibliography{my_bib}
\bibliographystyle{acl_natbib}

\newpage

\appendix

\section{Appendices}
\label{sec:appendix}

\begin{figure*}[b!]

\centering
\begin{tabular}{|l|l|l|} 
\hline
\textbf{Entity Type (SP)} & \textbf{Entity Type (TL)} & \textbf{Notes}                                         \\ 
\hline
Material                 & Material                  &                                                        \\ 
\hline
Number                   & Descriptor                &                                                        \\ 
\hline
Operation                & Operation                 &                                                        \\ 
\hline
Amount-Unit              & Descriptor                &                                                        \\ 
\hline
Condition-Unit           & Operation-Descriptor      &                                                        \\ 
\hline
Material-Descriptor      & Material-Descriptor       &                                                        \\ 
\hline
Condition-Misc           & Operation-Descriptor      &                                                        \\ 
\hline
Synthesis-Apparatus      & Synthesis-Apparatus       &                                                        \\ 
\hline
Nonrecipe-Material       & Null                      & Currently ignored, not part of synthesis      \\ 
\hline
Brand                    & Descriptor                &                                                        \\ 
\hline
Apparatus-Descriptor       & Synthesis-Apparatus-Descriptor                      &       \\ 
\hline
-                        & Mixture                   & Internal entity, represents a mixture  \\
\hline

                        &                    &   \\

\hline
\textbf{Relation Type (SP)} & \textbf{Action (TL)} &        \\
\hline
Participant-Material & input-assign &        \\
\hline
Apparatus-of & locate &        \\
\hline
Recipe-Target & obtain &        \\
\hline
Descriptor-of & link-descriptor &        \\
\hline
- & run-op &  Internal, used for simulating actions      \\
\hline
- & take/drop/examine & Native TextWorld actions on entities        \\
\hline
\end{tabular}
\caption{Central entity/relation types from the Synthesis Project schema (``SP''), and the corresponding \tlabs version (``TL'').}
\label{tab:ents}

\end{figure*}

\subsection{Entity \& Relation Types}
\label{subsec:types}

We have claimed that converting an annotation schema to a game for \tlabs was relatively straightforward. In this section,  we provide details of the mapping between the Synthesis Project annotation schema of (denoted with ``SP'' in the tables) to the \tlabs implementation (denoted ``TL''). A mapping between the central entity types is presented in Figure \ref{tab:ents}, as well as the \tlabs actions and representative corresponding relations in the schema. %
All current \tlabs entities and actions are shown here, though not all of the original entities and relations are listed. For the full mapping, refer to the project source repository.

\subsection{Synthetic Action-Graphs}\label{subsec:quests-dets}
\remove{\dnote{flow text. what's the purpose of this section}}

\begin{figure*}
\includegraphics[width=1.05\textwidth]{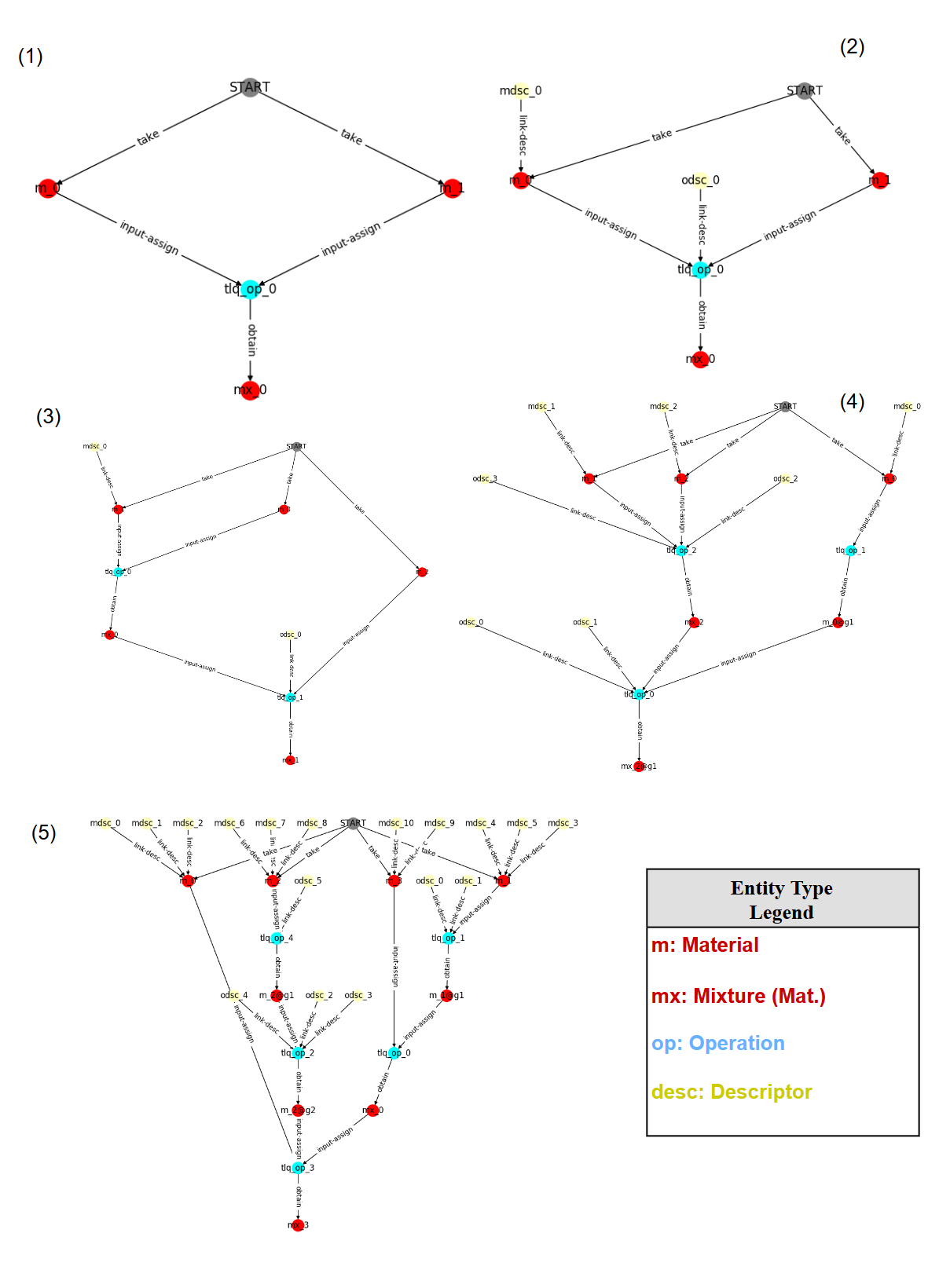}
\caption{\label{fig:quests}  Sample representative generated quests for various difficulty levels (listed in parentheses by  each graph). Each edge corresponds to an action in the text-based game.}
\end{figure*}

Figure \ref{fig:quests} displays sample representative generated quests for the various difficulty levels evaluated in Sec. \ref{sec:evaluation}, demonstrating the controllable complexity. As can be seen by comparison with the real text in Fig. \ref{fig:ttg} (which is only one sentence), these graphs correspond to short real-world surfaces, where even the hardest could by covered by a 2-3 sentence-long procedure.

\remove{
\begin{table}[]
\begin{tabular}{|l|l|}
\hline
Difficulty Level & Mean Quest Length (\# actions) \\ \hline
1                & 6                              \\
2                & 9.2                            \\
3                & 15.8                           \\
4                & 28.1                           \\
5                & 52.2                           \\ \hline
\end{tabular}
\caption{Avg. number of actions in action-graphs, by difficulty level.}
\label{tab:lengths}
\end{table}
}

\subsection{Action-Graphs from Real Annotated Graphs}\label{subsec:real-graphs}
\begin{figure*}
\includegraphics[width=1.00\textwidth]{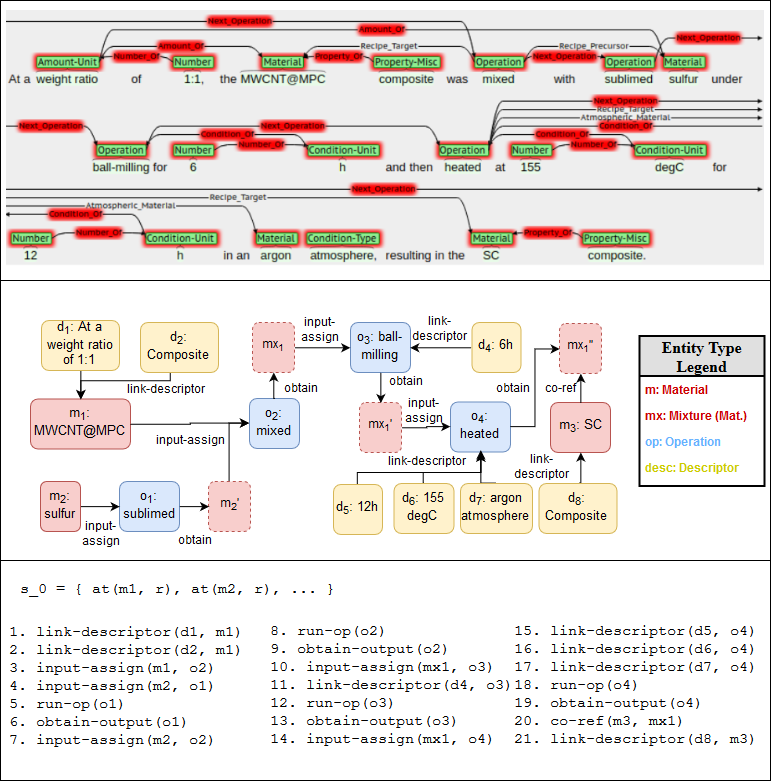}
\caption{\label{fig:ttg} Comparisons of the equivalent action graph representations. \textbf{Top}: Action graph section from Synthesis Project \citep{MSPT}. \textbf{Center}: \tlabs, showing same section with $K$ in graph form. Dashed borders indicate operation result entities which may be implicit in the text. \textbf{Bottom}: \tlabs with same $K$ as list of actions from initial state $s_0$.  }
\end{figure*}

We now provide further details on how the original Synthesis Project (SP) annotated graphs can be converted to a \tlabs action graph $K$. There are some minor differences between the formats, primarily in the handling of the SP ``next-operation'' relation. Rather than use a ``next-operation'' relation, we currently opt to explicitly model inputs/outputs to operations, as can be seen in Fig. \ref{fig:ttg}. This is a natural abstraction away from the surface text enabled by the grounded environment, and helps in tracking which materials participated in each operation, which is useful information for later analysis. Also, as noted, we currently use a simplified mapping (for example, many descriptor annotations such ``Amount-Unit'', ``Property-Misc'', etc. are chunked together as generic descriptors). In Fig. \ref{fig:ttg} we show $K$ both in action graph and action sequence form to demonstrate the equivalence. Also, we note that the ``next-operation'' annotations in \citet{MSPT} are currently just placeholders and not the true labels. For the purpose of demonstration, in Fig. \ref{fig:ttg} we manually add the correct annotation to our example (center and bottom).

\end{document}